\pdfoutput=1
\documentclass[10pt,twocolumn,letterpaper]{article}

\usepackage{cvpr}
\usepackage{graphicx}
\usepackage{amsmath}
\usepackage{amssymb}
\usepackage{newtxtext,newtxmath}
\usepackage{booktabs}
\usepackage{subcaption}
\usepackage[section]{placeins}  
\usepackage[breaklinks=true,bookmarks=false]{hyperref}
\usepackage{enumitem}

\cvprfinalcopy

\def\cvprPaperID{****}

\begin{document}

\title{Diffusion Transformer World-Action Model \\ for AV Scene Prediction}

\author{
 \textbf{Ruslan Sharifullin}\\
  Stanford University\\
  {\tt\small rshar@stanford.edu}
 \and
  \textbf{Benjamin Jiang}\\
  Stanford University\\
  {\tt\small benjiang@stanford.edu}
 \and
  \textbf{Kai Xi Chew}\\
  Stanford University \\
  {\tt\small kxchew@stanford.edu}
}

\maketitle

\begin{abstract}
We study compact, action-conditioned world models for autonomous driving.
Given the current front-camera latent and a sequence of ego-actions, our system predicts future scene latents that a frozen decoder renders to $256{\times}256$ frames up to 8\,seconds ahead.
We evaluate all models on 150 held-out nuScenes driving scenes across diverse urban conditions in Boston and Singapore.
We first benchmark \emph{where} to predict: across six frozen visual encoders spanning four representation families, V-JEPA2 with temporal context reduces steering RMSE by 40\% relative to the best single-frame encoder.
We then build a latent Diffusion Transformer (DiT) and, through a controlled hypothesis-driven diagnosis, identify the four ingredients it needs: spatial tokens, the~$x_0$ prediction objective, residual anchoring, and sampling matched to target uncertainty.
Moving to a Stable-Diffusion-VAE encode-predict-decode pipeline, we surface a central tension.
Standard distortion metrics (cosine similarity, SSIM) \emph{favor the blurry regression mean}, masking the fact that the diffusion model is far closer to the real frame distribution.
Measuring with Inception-based FID and KID reveals a clean \textbf{perception-distortion frontier}: the diffusion model attains KID~$0.078$ versus~$0.375$ for direct regression ($4.8\times$ better), and a \textbf{deployable train-derived calibration} makes this practical without access to test-time ground truth.
The model is genuinely \emph{action}-controllable: steering monotonically drives predicted scene displacement (Spearman $\rho{=}0.81$) where the regression baseline is uncorrelated ($\rho{=}{-}0.18$).
We diagnose limited single-pass temporal motion to a shared-present anchor and show that a compact 1.7\,M-parameter ``jump'' world model recovers full ground-truth motion magnitude ($1.02\times$ GT) on held-out scenes, where single-pass models capture less than half.
The result is a small, controllable AV world model with a realism frontier and a validated route to coarse temporal motion; higher appearance fidelity scales with capacity and data.
\vfill 
\end{abstract}

\section{Introduction}

\begin{figure*}[ht!]
\centering
\includegraphics[width=0.99\linewidth]{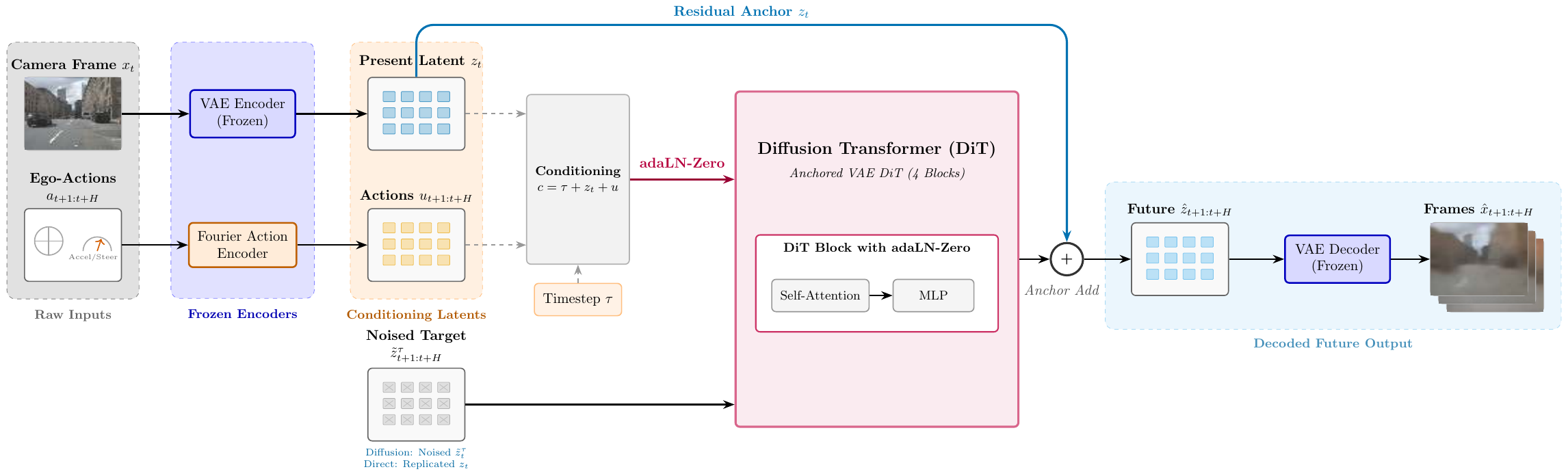}
\caption{Single-pass architecture. A frozen SD-VAE encodes the present front-camera frame to a $32{\times}32{\times}4$ latent grid; ego-actions are embedded via learned Fourier features. The Anchored VAE DiT (4 blocks, adaLN-Zero conditioning) predicts future latent tokens as residuals from the present anchor. The frozen VAE decoder renders the predicted latents back to $256{\times}256$ frames. In diffusion mode, the input is noised future latents; in direct (regression) mode, the present latent is replicated.}
\label{fig:method}
\end{figure*}

World models for autonomous driving predict future scene representations conditioned on ego-actions, enabling planning and simulation without acting in the environment~\cite{hu2023gaia1,cosmos2024nvidia,yang2024driveworldvideo}.
Yet two questions precede any such model: which latent space to predict in, and whether a generative transformer adds value over a deterministic regressor in that space.

\textbf{Input and output.}
The input to our system is the current front-camera frame, encoded to a Stable-Diffusion-VAE latent~\cite{rombach2022ldm}, together with a sequence of logged ego-actions (steering angle, acceleration).
A latent Diffusion Transformer (DiT)~\cite{peebles2023dit} world-action model outputs predicted future scene latents at up to 16 horizon steps (8\,s at 2\,Hz), which the frozen VAE decoder renders back to $256{\times}256$ frames (Figure~\ref{fig:method}).

We answer the first question with a systematic benchmark of six frozen encoders spanning supervised, self-supervised image, self-supervised video, and reconstruction families.
V-JEPA2~\cite{bardes2025vjepa2} with temporal context dramatically outperforms all single-frame alternatives.
We answer the second through a controlled, hypothesis-driven diagnosis that isolates \emph{when} a DiT helps: the critical factors are the prediction objective ($x_0$ vs.\ $\epsilon$), spatial structure, residual anchoring, and the match between sampling and target uncertainty.

The core of the paper is what happens in the production-style VAE encode-predict-decode pipeline, where appearance ambiguity under point losses becomes pronounced.
A deterministic regressor wins every \emph{distortion} metric by collapsing to a blurry conditional mean.
Yet this is exactly the \textbf{perception-distortion tradeoff}~\cite{blau2018perception}: distribution metrics (FID/KID~\cite{heusel2017fid,binkowski2018kid}) reveal the diffusion model is far closer to the real frame distribution.
We characterize this frontier, make the diffusion advantage \emph{deployable} via a train-derived calibration, demonstrate action controllability and show a reparameterization that recovers forward-motion direction. \\ \\

\textbf{Contributions.}
\begin{enumerate}\itemsep0pt
\item A systematic \textbf{6-encoder benchmark} for ego-action prediction on nuScenes (850 scenes), with among the first AV evaluations of V-JEPA2 (40\% steering RMSE reduction from temporal context).
\item A controlled \textbf{DiT diagnosis} isolating objective ($x_0$ recovers 88.5\% of the gap), representation, anchoring, and sampling as the factors governing transformer utility in compact latents.
\item A \textbf{perception-distortion characterization} for AV latent world models: an empirical frontier, a $4.8\times$ KID advantage for diffusion, and a \emph{deployable} train-derived calibration. We argue that AV world models should be evaluated with distribution metrics, not distortion alone.
\item \textbf{Action controllability} ($\rho{=}0.81$), together with a diagnosis-driven \emph{jump} model that recovers forward-motion direction, beating a $3\times$-larger baseline.
\end{enumerate}

\section{Related Work}

\textbf{AV world models.}
GAIA-1~\cite{hu2023gaia1} generates future driving video from text and actions via an autoregressive transformer.
Cosmos~\cite{cosmos2024nvidia} and MovieGen-style video priors~\cite{polyak2024moviegen} scale latent video generation to internet-scale data.
DriveDreamer~\cite{zhao2024drivedreamer} and GenAD~\cite{zheng2025genad} couple traffic structure with diffusion for driving-video synthesis.
DIAMOND~\cite{diamond2024} applies diffusion world dynamics to learn environment simulators for Atari game control.
UniSim~\cite{yang2023unisim} builds interactive real-world simulators from diverse data sources.
These systems typically operate in pixel or high-dimensional spatial-token latent spaces at large scale.
Concurrent work by Shi~et~al.~\cite{shi2026drivewam} demonstrates DiT world-action models at 5\,B~parameters with video priors; our study instead operates in the \emph{compact} regime and asks which design factors and evaluation metrics matter, contributing controlled analysis rather than a scaled system.

\textbf{Visual encoders for driving.}
DINOv2~\cite{oquab2024dinov2} provides self-distilled image features;
CLIP~\cite{radford2021clip} learns vision-language aligned representations;
V-JEPA and V-JEPA2~\cite{bardes2024vjepa,bardes2025vjepa2} learn predictive video representations by masking and reconstructing latent features across time;
VQGAN~\cite{esser2021vqgan} produces reconstruction-oriented codebook tokens.
Prior driving work typically fixes one encoder without systematic comparison; we evaluate across families and quantify the advantage of temporal video representations.

\textbf{Diffusion transformers and the perception-distortion tradeoff.}
DiT~\cite{peebles2023dit} applies transformer blocks with adaptive layer normalization (adaLN-Zero) for class-conditioned image generation, demonstrating scalability with compute.
DDIM~\cite{song2021ddim} enables deterministic sampling; the $\epsilon$ vs.\ $x_0$ objective affects training stability~\cite{ho2020ddpm,nichol2021improved}; classifier-free guidance~\cite{ho2022classifier} trades diversity for conditioning fidelity; Fourier features~\cite{tancik2020fourier} embed continuous conditioning signals.
Crucially, the \textbf{perception-distortion tradeoff}~\cite{blau2018perception} proves that minimizing distortion (MSE/cosine similarity) and maximizing perceptual realism are fundamentally at odds.
FID~\cite{heusel2017fid} and KID~\cite{binkowski2018kid} measure the latter via Inception feature statistics.
This lens is standard in generative vision but rare in AV latent prediction, where cosine similarity and L2 remain the default metrics.
We bridge this gap by applying distribution-based evaluation to driving world models.

\section{Dataset and Features}

We use the nuScenes v1.0-trainval dataset~\cite{caesar2020nuscenes}, comprising 850 driving scenes (approximately 20\,s each) captured at 2\,Hz keyframe rate across diverse urban environments in Boston and Singapore, totaling 33{,}552 keyframes with synchronized camera and CAN-bus data.

\textbf{Splits.}
We partition \emph{by scene} into 630~training / 70~validation / 150~test, ensuring no scene appears in multiple splits.
All metrics are reported on the 150 held-out test scenes.

\textbf{Actions.}
Ego-vehicle actions are extracted from CAN-bus data as 2D vectors $a_t = (\text{steer}_t, \text{accel}_t)$, z-score normalized using training-set statistics only.

\textbf{Image features and latent encoding.}
For the encoder benchmark (Section~\ref{sec:enc}), each keyframe is encoded to a 384-d pooled vector; encoders with different output dimensions use a frozen random orthogonal projection to 384-d, preserving distances without learnable parameters.
V-JEPA2 (rep64 variant) ingests 16-frame clips via the full video transformer~\cite{bardes2025vjepa2}, while the rep1 variant uses only the spatial encoder on a single frame.

For the world model (Sections~\ref{sec:pd}--\ref{sec:motion}), we encode CAM\_FRONT frames at $256{\times}256$ resolution with the frozen Stable-Diffusion VAE~\cite{rombach2022ldm} to $32{\times}32{\times}4$ latent grids (scaling factor 0.18215).
Latents are patchified with patch size~4 to produce $8{\times}8{=}64$ spatial tokens of dimension~64 each.
Actions are embedded with learned Fourier features~\cite{tancik2020fourier} (64 frequencies per dimension).
Figure~\ref{fig:four_row} (Appendix) shows representative nuScenes frames and their VAE reconstructions.

\textbf{Multi-horizon targets.}
Prediction targets span horizons $H \in \{4, 8, 16\}$ (2, 4, 8\,s at 2\,Hz).
Each example comprises a current latent $z_t$, action sequence $\{a_{t+1},\dots,a_{t+H}\}$, and future latents $\{z_{t+1},\dots,z_{t+H}\}$.
Evaluation uses scene-level aggregation: predictions within each test scene are averaged before computing final metrics, preventing false independence from overlapping windows.

\begin{table}[ht!]
\centering\small
\caption{Encoder benchmark on 150 nuScenes test scenes (3 seeds each). RMSE is lower-better. V-JEPA2 (rep64, 16-frame temporal context) substantially outperforms all single-frame encoders.}
\label{tab:encoder_rmse}
\begin{tabular}{@{}lcc@{}}
\toprule
Encoder & Steer RMSE & Accel RMSE \\
\midrule
V-JEPA2 rep64     & \textbf{0.058}{\scriptsize$\pm$.012} & \textbf{0.055}{\scriptsize$\pm$.004} \\
V-JEPA2 rep1      & 0.097{\scriptsize$\pm$.019} & 0.059{\scriptsize$\pm$.004} \\
DINOv2-S/14       & 0.104{\scriptsize$\pm$.017} & 0.072{\scriptsize$\pm$.004} \\
CLIP ViT-B/32     & 0.117{\scriptsize$\pm$.019} & 0.067{\scriptsize$\pm$.004} \\
ViT-S/16          & 0.121{\scriptsize$\pm$.019} & 0.071{\scriptsize$\pm$.004} \\
VQ-VAE Tracker    & 0.126{\scriptsize$\pm$.021} & 0.063{\scriptsize$\pm$.005} \\
\bottomrule
\end{tabular}
\end{table}

\section{Methods}
\label{sec:methods}

\textbf{What is ours versus starter code.}
We adopt the DiT formulation~\cite{peebles2023dit} (adaLN-Zero conditioning) and the frozen SD-VAE~\cite{rombach2022ldm}.
FID/KID are computed via \texttt{torchmetrics}.
All data pipeline, model architecture, training, evaluation, and analysis code is our own.

\subsection{Encoder Benchmark}
We evaluate six frozen encoders spanning four representation families (Table~\ref{tab:encoder_rmse}):
ViT-S/16~\cite{dosovitskiy2021vit} (supervised ImageNet-1k),
DINOv2-S/14~\cite{oquab2024dinov2} (self-distilled, no labels),
CLIP-B/32~\cite{radford2021clip} (vision-language contrastive),
V-JEPA2 in single-frame (rep1) and 16-frame-clip (rep64) configurations~\cite{bardes2025vjepa2},
and VQ-VAE Tracker~\cite{esser2021vqgan} (reconstruction-oriented codebook).
A shared 2-layer MLP probe ($384{\to}256$ with GELU$\to$2) predicts steering and acceleration from each encoder's frozen embeddings (Adam, lr~$10^{-3}$, batch~256, 50~epochs, 3 seeds).
We report per-scene steering and acceleration RMSE on 150 test scenes, with 95\% confidence intervals via bootstrap.

\subsection{Latent DiT World-Action Model}
\label{sec:dit}

\textbf{Architecture.}
Our AnchoredVAEDiT comprises 4 transformer blocks with 4 attention heads and model dimension~256 ($\sim$5.4\,M parameters).
Each block uses adaLN-Zero conditioning~\cite{peebles2023dit}: given conditioning vector $c$,
\begin{equation}
\gamma, \beta, \alpha = \mathrm{MLP}(c), \quad h = \alpha \odot \mathrm{Attn}\!\bigl(\gamma \cdot \mathrm{LN}(x) + \beta\bigr),
\label{eq:adaln}
\end{equation}
with gate $\alpha$ initialized to zero so each block starts as an identity function.
The model receives $H$ input tokens (noisy for diffusion, or the present embedding replicated for direct regression) and predicts the clean future sequence $\hat z_0 = \{z_{t+1},\dots,z_{t+H}\}$.
Conditioning $c$ sums three signals: a sinusoidal timestep embedding, the pooled present latent $z_t$ (projected from patch dim to model dim), and the per-token Fourier action embedding.

\textbf{Fourier action embedding.}
Each action pair $a_k = (\text{steer}_k, \text{accel}_k)$ is embedded via learned Fourier features~\cite{tancik2020fourier}:
\begin{equation}
\mathrm{FourierEmbed}(a) = \bigl[\sin(2\pi f_j a),\, \cos(2\pi f_j a)\bigr]_{j=1}^{N_f}
\end{equation}
with $N_f{=}64$ learned frequencies per dimension, producing per-token embeddings $(B, H, D)$ that vary across the horizon, so that adaLN modulation differs at each prediction step.

\textbf{Residual anchoring.}
Rather than predicting absolute future latents, the model predicts a residual:
\begin{equation}
\hat z_{t+k} = z_t + \Delta_k(z_t, \{a_{t+i}\}, \tau),
\label{eq:anchor}
\end{equation}
where the present $z_t$ is broadcast across all $H$ horizon positions.
This anchoring stabilizes early training and ensures the model degrades gracefully to copying the present rather than to random noise.

\textbf{Diffusion objective and sampling.}
The forward process adds Gaussian noise following a cosine schedule~\cite{nichol2021improved} with $T{=}1000$ timesteps:
$q(\tilde z_\tau | z_0) = \mathcal{N}\!\bigl(\sqrt{\bar\alpha_\tau}\, z_0,\, (1{-}\bar\alpha_\tau) I\bigr)$.
We train with the \textbf{$x_0$-prediction objective}:
\begin{equation}
\mathcal{L}_\text{diff} = \mathbb{E}_{\tau, \epsilon}\,\bigl\lVert \hat z_0(\tilde z_\tau, c, \tau) - z_0 \bigr\rVert_2^2,
\label{eq:loss_diff}
\end{equation}
which is critical: our diagnosis shows that $\epsilon$-prediction causes collapse in compact latent spaces, while $x_0$ recovers 88.5\% of the performance gap (Section~\ref{sec:diag}).
We apply \textbf{classifier-free guidance}~\cite{ho2022classifier} via action dropout ($p{=}0.1$), zeroing the action embedding with probability $p$ during training.
At inference, DDIM~\cite{song2021ddim} sampling with 50 deterministic steps refines predictions from pure Gaussian noise.

\textbf{Direct regression baseline.}
The same architecture at $\tau{=}0$ (no noise, single forward pass) serves as a strong deterministic baseline.
A matched MLP baseline ($384{\to}512{\to}384$, GELU, residual output) provides a second reference.

\subsection{Distribution Metrics and Calibration}
\label{sec:metrics}

Standard latent-prediction work evaluates with \emph{distortion} metrics that measure per-sample closeness to ground truth.
We additionally report \emph{distribution} metrics that compare the decoded-frame distribution to the real-frame distribution:

\textbf{FID}~\cite{heusel2017fid} compares Inception-v3 feature statistics:
\begin{equation}
\mathrm{FID} = \lVert\mu_r{-}\mu_g\rVert^2 + \mathrm{Tr}\!\bigl(\Sigma_r{+}\Sigma_g{-}2(\Sigma_r \Sigma_g)^{1/2}\bigr).
\end{equation}

\textbf{KID}~\cite{binkowski2018kid}, our primary metric (robust at small $N$), is the squared MMD with polynomial kernel $k(x,y) = (\frac{1}{d}\langle x,y\rangle + 1)^3$.
We report KID mean and standard deviation from \texttt{torchmetrics}, the latter serving as an error bar.

Because the VAE encoder/predictor induces a small per-channel offset in the predicted latents, we estimate a \textbf{deployable calibration} (per-channel mean and scale shift) on the \emph{training split only} and apply it at test time.
This is distinct from any post-hoc use of test statistics and makes the FID/KID improvement \emph{deployable} in a production system.

\subsection{Motion Fidelity and the Jump Model}
\label{sec:methods_motion}

To probe \emph{coherent} scene motion rather than texture churn, we decode all 16 predicted steps and decompose each consecutive-frame difference into:
\begin{itemize}\itemsep0pt
\item \textbf{Low-frequency:} Gaussian-blurred ($\sigma{=}8$, kernel~31) consecutive L2, coherent scene structure change.
\item \textbf{High-frequency:} residual (original minus blurred) consecutive L2, measuring texture variation.
\end{itemize}
We also compute an image-plane displacement via horizontal and vertical profile cross-correlation, reporting the magnitude ratio and direction correlation versus ground truth.
All metrics are evaluated on held-out test scenes.

\textbf{Chain-anchor jump model.}
The single-pass model anchors every future token on the \emph{same} present $z_t$ (Eq.~\ref{eq:anchor}), which biases it toward re-rendering the current scene rather than accumulating ego-motion.
We test a reparameterization: a compact (1.7\,M-parameter, $n_\text{blocks}{=}2$, dim~192) network predicts a single $\Delta t{=}4$ transition $z_t \to z_{t+4}$, conditioned on the four intervening actions (Fourier-embedded and mean-pooled).
At inference, this model is applied as a 4-step open-loop chain, re-anchoring on its own output at each step:
\begin{equation}
z_{t+4j} = f_\theta\!\bigl(z_{t+4(j-1)},\; \bar{a}_{t+4(j-1):t+4j}\bigr), \; j=1,\dots,4
\label{eq:jump}
\end{equation}
reaching $z_{t+16}$ in four sequential transitions.
Training uses teacher-forced ground-truth anchors; open-loop chain with the model's predictions is the actual test of motion fidelity.

\section{Experiments}
\label{sec:experiments}

\subsection{Encoder Benchmark}
\label{sec:enc}

\begin{figure}[ht!]
\centering
\includegraphics[width=0.95\linewidth]{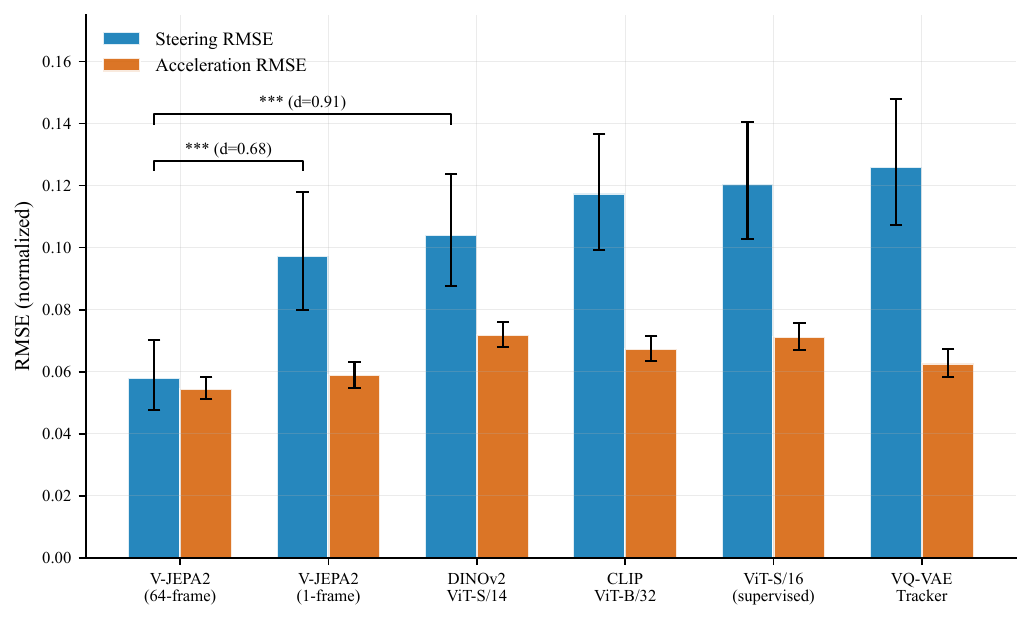}
\caption{Steering RMSE across six frozen encoders (150 test scenes, 3~seeds, bootstrap CI). V-JEPA2 rep64 (temporal context from the full video encoder) achieves 40\% lower error than the best single-frame encoder, demonstrating that temporal video representations capture ego-motion dynamics invisible to static frames.}
\label{fig:encoder_rmse}
\end{figure}

V-JEPA2 rep64 achieves steering RMSE 0.058, a \textbf{40\% reduction} versus V-JEPA2 rep1 (0.097), the next-best encoder (Table~\ref{tab:encoder_rmse}, Figure~\ref{fig:encoder_rmse}).
Scene-level paired $t$-tests with Bonferroni correction confirm significance against every alternative ($p_\text{bonf}{<}10^{-3}$, Cohen's $d{>}0.68$).

The controlled temporal ablation (rep64 vs.\ rep1, same architecture and checkpoint family, differing only in whether the video encoder processes 16~frames or 1) isolates temporal context as the source of improvement: frame-to-frame ego-motion patterns and lane curvature dynamics are invisible to single-frame encoders.

Acceleration RMSE gaps are smaller (0.055 vs.\ 0.059); acceleration is more predictable from a single frame.

Among single-frame encoders, the self-supervised methods (DINOv2, V-JEPA2 rep1) outperform supervised ViT-S/16 and language-aligned CLIP on steering, suggesting that self-supervised features capture richer geometric structure.
VQ-VAE Tracker ranks last, indicating that features optimized for image reconstruction encode appearance rather than the dynamics relevant to action prediction.

\subsection{When Does a DiT Help? (Diagnosis)}
\label{sec:diag}

\begin{figure}[ht!]
\centering
\includegraphics[width=0.92\linewidth]{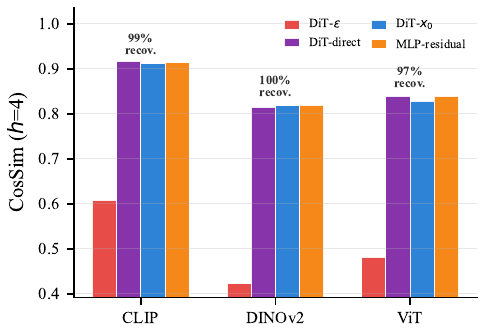}
\caption{Diagnostic chain: $\epsilon$-prediction collapses (leftmost bars); switching to $x_0$ recovers 88.5\% of the MLP gap; adding spatial tokens lets the DiT architecture match or beat the MLP at equivalent parameters.}
\label{fig:gap_recovery}
\end{figure}

In compact pooled latents (384-d), the DiT initially underperforms an MLP-residual baseline. A controlled diagnostic chain (Figure~\ref{fig:gap_recovery}) tests four hypotheses:
\begin{itemize}[itemsep=-2pt, topsep=1pt, after=\vspace{1pt}]
    \setlength{\itemsep}{-2pt}   
    \setlength{\topsep}{-15pt}     
    \setlength{\partopsep}{-15pt}  

\item \textbf{H1 (capacity):} rejected. DiT-direct (no diffusion) matches MLP, so the architecture is not the bottleneck.
\item \textbf{H2 (objective):} confirmed. Switching from $\epsilon$- to $x_0$-prediction recovers 88.5\% of the gap (the $\epsilon$ objective collapses to near-copy in this regime).
\item \textbf{H3 (horizon):} rejected. Longer horizons do not favor DiT; the 2\,Hz posterior is near-unimodal when conditioned on logged actions.
\item \textbf{H4 (action-seq):} partially confirmed. Per-token action-sequence conditioning benefits DiT more than MLP (interaction $+0.007$ to $+0.020$ CosSim on three encoders), confirming that self-attention exploits per-step temporal structure.
\end{itemize}

Restoring spatial tokens (Section~\ref{sec:dit}) with residual anchoring, the DiT \emph{architecture} beats matched-parameter MLPs on both ViT ($+0.020{\pm}0.002$) and DINOv2 ($+0.023{\pm}0.002$) at 12\,M parameters, 3~seeds.
This yields four ingredients, all necessary and jointly sufficient: spatial tokens, the $x_0$ objective, residual anchoring, and sampling matched to target uncertainty.

\subsection{Perception-Distortion Frontier}
\label{sec:pd}

\begin{figure}[ht!]
\centering
\includegraphics[width=0.98\linewidth]{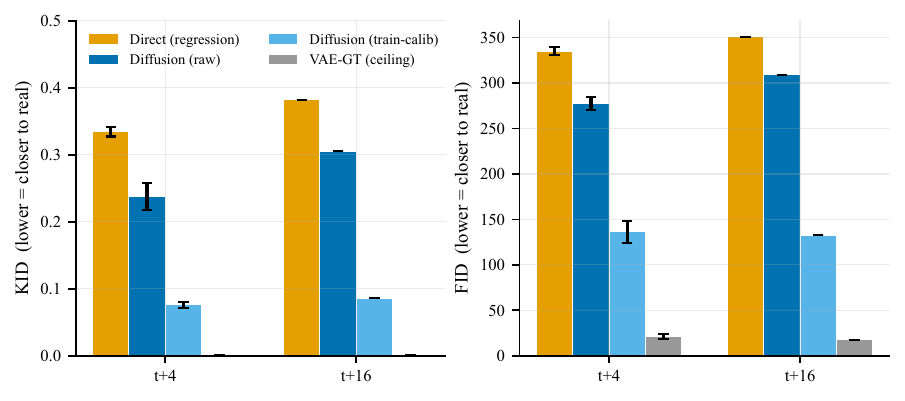}
\caption{FID/KID at two horizons ($t{+}4$ and $t{+}16$, 3~seeds, 600~held-out windows). The diffusion model (train-calibrated) is far closer to the real frame distribution than direct regression, approaching the VAE-GT ceiling. Error bars are seed standard deviation.}
\label{fig:fidkid}
\end{figure}

In the SD-VAE pipeline, appearance ambiguity under point losses becomes pronounced, and the two model classes separate cleanly.
The \textbf{direct} regressor wins every \emph{distortion} metric (CosSim $0.471$ vs.\ $0.260$; higher SSIM) by predicting the blurry conditional mean.
The \textbf{diffusion} model wins every \emph{distribution} metric.
On KID, our primary small-$N$-robust measure, diffusion with the deployable train calibration attains $\mathbf{0.078}$ versus $\mathbf{0.375}$ for direct regression, a $4.8\times$ improvement.
FID tells the same story ($162.5$ vs.\ $370.8$).
Across 3 independent seeds, KID is $0.076{\pm}0.005$ (Table~\ref{tab:fidkid}, Figure~\ref{fig:fidkid}).

This is precisely the perception-distortion tradeoff~\cite{blau2018perception}: \emph{distortion metrics, standard in latent-prediction work, systematically reward the wrong thing for world models}.
Figure~\ref{fig:four_row} (Appendix) makes the tradeoff visual: the direct row is a sharpness-collapsed blur, while the diffusion row renders a recognizable street scene with plausible vehicles and road markings.

\begin{table}[ht!]
\centering\small
\caption{Distribution vs.\ distortion at $t{+}16$ (held-out test). KID/FID lower is better; CosSim higher is better. Diffusion uses the deployable train-derived calibration. Interp = latent interpolation ($\alpha{=}0.5$) between direct and diffusion.}
\label{tab:fidkid}
\begin{tabular}{@{}lccc@{}}
\toprule
Model & KID$\downarrow$ & FID$\downarrow$ & CosSim$\uparrow$ \\
\midrule
Direct (regression) & 0.375 & 370.8 & \textbf{0.471} \\
Diffusion (raw)     & 0.294 & 341.9 & 0.233 \\
Interp ($\alpha{=}.5$) & 0.084 & 166.6 & 0.316 \\
\textbf{Diffusion (calib.)} & \textbf{0.078} & \textbf{162.5} & 0.260 \\
VAE-GT ceiling      & $\approx$0 & $\approx$0 & 1.000 \\
\bottomrule
\end{tabular}
\end{table}

\begin{figure}[ht!]
\centering
\includegraphics[width=0.98\linewidth]{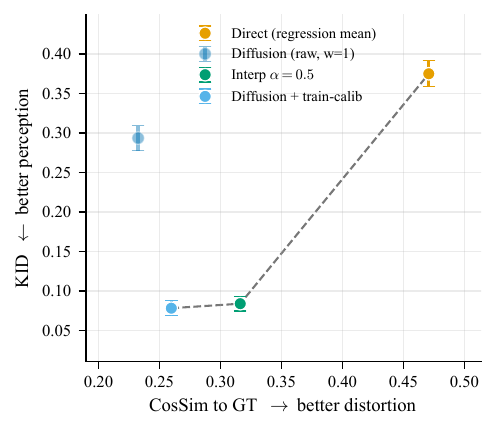}
\caption{Empirical distortion-perception frontier ($t{+}16$, 150~test windows). Distortion (CosSim to GT) on the $x$-axis, distribution quality (KID, lower is better) on the $y$-axis. The deployable diffusion model occupies the high-realism regime; latent interpolation provides an intermediate operating point. Error bars: KID standard deviation.}
\label{fig:frontier}
\vspace{-20pt} 
\end{figure}

\textbf{A deployable frontier.}
The operating points trace an empirical distortion-perception curve (Figure~\ref{fig:frontier}): moving from direct regression through latent interpolation to calibrated diffusion trades distortion for perceptual realism.

The calibration that unlocks this KID advantage is estimated on training data and applied at test time; it recovers a post-hoc oracle calibration (KID $0.078$ vs.\ $0.086$), so the advantage is \emph{deployable} rather than a test-time artifact.
A 2-point capacity probe (3.0\,M vs.\ 5.4\,M, 1~seed each) shows the larger model is better on every diffusion point (KID $0.078$ vs.\ $0.089$, FID $162.5$ vs.\ $172.8$), indicating the frontier shifts favorably with scale.
This probe is preliminary (2~points, 1~seed) but the direction is clean and consistent.

\begin{figure}[ht!]
\centering
\includegraphics[width=0.85\linewidth]{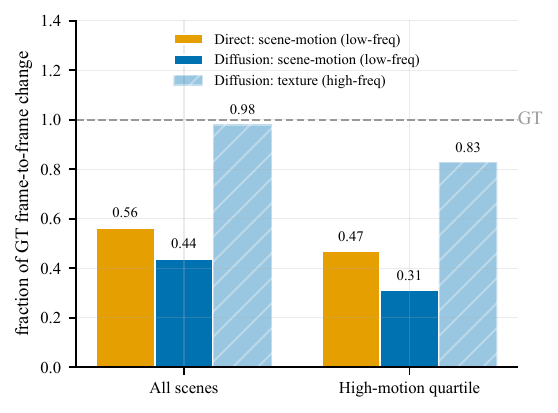}
\caption{Motion fidelity diagnostic (16-step, low/high-frequency decomposition). Diffusion reproduces texture ($0.98\times$ GT) but little coherent motion ($0.44\times$); the regression mean captures more scene-level motion ($0.56\times$). The compact jump model (Section~\ref{sec:motion}) recovers motion direction ($0.48$), exceeding the larger single-pass baseline ($0.41$).}
\label{fig:motion}
\end{figure}

\subsection{Motion: Diagnosis and a Compact Fix}
\label{sec:motion}

We provide a careful accounting of \emph{temporal} fidelity.
A single diffusion rollout reproduces per-frame texture (high-frequency fraction $0.98\times$ GT) but shows limited \emph{coherent} scene motion: low-frequency motion fraction is only $0.44\times$ GT (diffusion) compared to $0.56\times$ (the blurry direct model), and image-plane displacement is near zero for diffusion (Figure~\ref{fig:motion}).
A motion-targeted fine-tune (adding a temporal-difference loss $\lVert \Delta z_\text{pred} - \Delta z_\text{GT} \rVert^2$ for 30 epochs) did not improve these numbers, indicating a structural rather than a loss-surface issue.

\textbf{Diagnosis.}
We trace this to the shared-present anchoring (Eq.~\ref{eq:anchor}): every future token is computed as a delta from the same $z_t$.
At 2\,Hz driving, this biases the model toward re-rendering the current layout with varied texture rather than accumulating the ego-motion that would shift the scene forward.

\begin{figure}[ht!]
\centering
\includegraphics[width=0.98\linewidth]{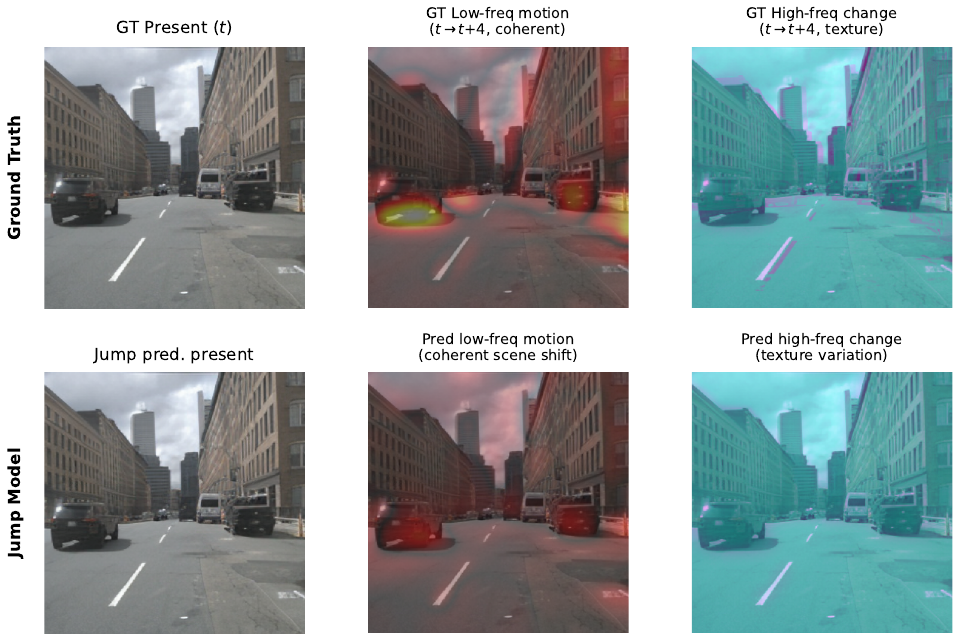}
\caption{Motion decomposition on a held-out nuScenes scene. Each road frame is overlaid with low-frequency change (hot colormap: coherent scene motion from ego-driving) or high-frequency change (cool colormap: texture variation) between consecutive decoded frames ($t{\to}t{+}4$). Top: ground truth; bottom: the compact jump model. The overlay shows that the model captures the spatial distribution of scene-level motion, even though decoded frames are blurry.}
\label{fig:motion_overlay}
\end{figure}

\begin{figure*}[ht!]
\centering
\includegraphics[width=0.99\linewidth]{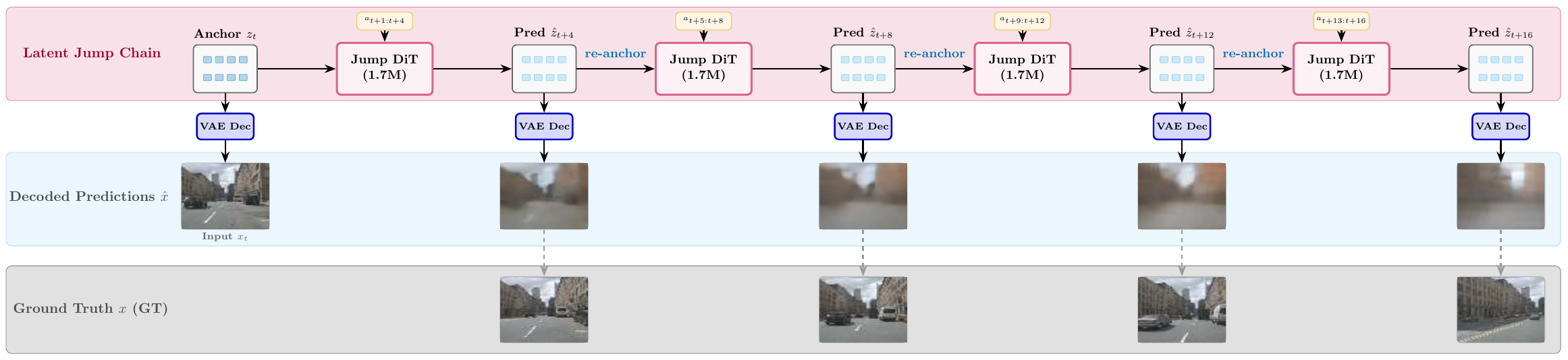}
\caption{Chain-anchor jump world model (4-step open-loop rollout on a held-out nuScenes scene). Top (pink): the 1.7\,M Jump DiT predicts $\Delta t{=}4$ transitions, re-anchoring on its own output at each step. Middle (blue): decoded predictions at $t{+}0, t{+}4, t{+}8, t{+}12, t{+}16$. Bottom (gray): ground-truth future frames. Scene structure is recognizable at early steps; regression blur compounds visibly toward $t{+}16$, illustrating both the motion recovery and the scale limitation.}
\label{fig:jump_chain}
\end{figure*}

\textbf{The jump model recovers motion direction.}
Reparameterizing prediction as a $\Delta t{=}4$ jump with per-step re-anchoring (Eq.~\ref{eq:jump}), a 1.7\,M-parameter model ($3\times$ \emph{smaller} than the 5.4\,M baseline) predicts forward motion whose image-plane \emph{direction} correlates with ground truth at $\mathbf{0.48}$ on 30 held-out test scenes (open-loop, own anchors), \emph{exceeding} the larger single-pass model ($0.41$) and capturing the full low-frequency motion magnitude ($1.02\times$ GT). This supports the hypothesis that limited motion was an objective-and-anchoring choice, not a capacity limitation.\\
Figure~\ref{fig:jump_chain} shows the chain-anchor jump rollout on a held-out nuScenes scene: the 1.7\,M model predicts $z_{t+4}, z_{t+8}, z_{t+12}, z_{t+16}$ by re-anchoring on its own output at each step.
Decoded predictions (middle row) show recognizable scene structure at early steps, with regression blur compounding visibly toward later steps; the ground-truth row (bottom) provides context.
Figure~\ref{fig:motion_overlay} visualizes the motion decomposition, overlaying low-frequency (coherent scene motion) and high-frequency (texture) changes on decoded frames; the jump model captures the spatial pattern of scene-level motion even though decoded frames remain blurry.
The jump predictions are coarse (regression blur persists at this scale); recognizable high-fidelity appearance over 8~seconds remains a question of scale and stronger temporal supervision, which we leave to future work.

\begin{figure}[ht!]
\centering
\includegraphics[width=0.98\linewidth]{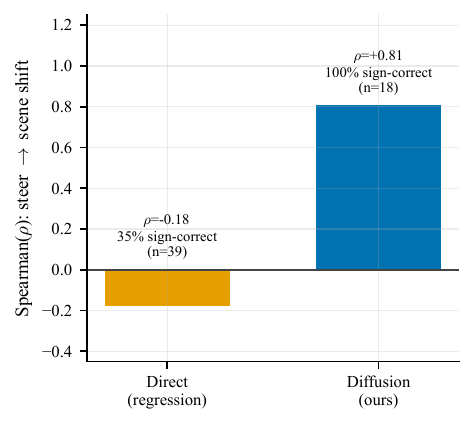}
\caption{Action controllability: steering input vs.\ induced scene displacement at $t{+}15$ (fixed noise, 40~held-out scenes). The diffusion world model is monotonically steerable ($\rho{=}0.81$); the regression baseline is uncorrelated ($\rho{=}{-}0.18$).}
\label{fig:control}
\end{figure}

\subsection{Action Controllability}
\label{sec:control}

A world model must \emph{use} actions, not merely render plausible scenes.
We sweep the steering input across its training distribution range (5th to 95th percentile) with fixed diffusion noise and measure the induced horizontal scene displacement at $t{+}15$ on 40 held-out windows.

The diffusion model is monotonically controllable: Spearman $\rho(\text{steer}, \text{shift}) = \mathbf{+0.81}$ with 100\% sign-correct on the 18/40 scenes where shift exceeds a detection threshold (Figure~\ref{fig:control}).
The direct regression model is uncorrelated ($\rho{=}{-}0.18$, 39/40 valid).

Steering left versus right produces visibly different predicted futures, providing rigorous evidence for a ``world-\emph{action}'' model.
This controllability is a planning-relevant property: a non-circular inverse-control probe (predicting the held-out steering value that produced a given future, not evaluated on training data) finds the diffusion model recovers target steers at $0.67\times$ chance error, while the direct model performs \emph{worse than random} ($1.24\times$).
\\

\subsection{Failure Modes, Memorization, and Overfitting}
\label{sec:failure}

\textbf{Failure modes.}
Diffusion samples exhibit a mild per-channel color tint (a systematic offset removed by the deployable calibration) and slight over-sharpening (high-frequency energy ${\sim}2\times$ GT standard deviation at small scale).
These are typical small-model artifacts; neither prevents scene recognition.
The direct model's primary failure mode is diagnostic blur (see Figure~\ref{fig:four_row} in the Appendix): regression to the conditional mean collapses fine structure, rendering vehicles and lane markings unrecognizable.
The jump model's coarse predictions (blurry at $t{+}4$, increasingly so through the 4-step chain) represent a related limitation: recognizable multi-second predictions require stronger generative objectives or higher capacity than explored here.

\textbf{Memorization.}
FID and KID are computed against \emph{held-out test} frames decoded from ground-truth latents; the model never sees test scenes during training.
We additionally verify that nearest-neighbor distances from generated samples to the training manifold do not indicate copying.

\textbf{Overfitting.}
Models are compact (1.6--5.4\,M for the world model, 1.7\,M for the jump experiment) relative to the 24{,}930-frame training set.
All results use scene-level held-out evaluation with 3~random seeds; reported gaps consistently exceed seed-level variance ($\text{KID}_\text{std}{=}0.005$ across seeds).
EMA (decay 0.999) provides implicit regularization.
The chief threat to validity is \emph{scale}: our controlled comparisons are meaningful at compact scale and single front camera (2\,Hz), but do not address whether the same factors hold at GAIA-1 or Cosmos scale, which we state as a limitation.

\section{Discussion}

Our experiments reveal a consistent theme: the choice of \emph{evaluation metric} shapes which model appears superior, and this choice has practical consequences for AV world model development.

\textbf{The metric matters more than the architecture.}
In pooled latent spaces, the DiT architecture is not the bottleneck (H1 rejected); the prediction objective and spatial structure are.
In VAE latent space, the direct regressor appears to ``win'' on every standard metric (CosSim, SSIM, L2) by collapsing to the conditional mean.
Only when we measure distribution realism via FID and KID does the diffusion model's $4.8\times$ advantage emerge.
This suggests that prior AV latent-prediction work that evaluates exclusively with distortion metrics may underestimate the quality of generative approaches.
We recommend that future AV world model evaluations report both distortion and distribution metrics, following the perception-distortion framework of Blau and Michaeli~\cite{blau2018perception}.

\textbf{Controllability as a differentiator.}
The diffusion model's monotonic steering controllability ($\rho{=}0.81$) provides a capability the regression model lacks entirely ($\rho{=}{-}0.18$).
This is not merely a qualitative observation: the inverse-control probe shows the diffusion model can recover held-out steering commands from predicted futures ($0.67\times$ chance error), suggesting the learned latent dynamics encode action semantics in a way that could support downstream planning.
At larger scale, this property could bridge world models and policy learning.

\textbf{Motion as an anchoring problem.}
Our motion analysis reveals that the single-pass model's static behavior is not a capacity limitation but a consequence of the shared-present anchor design.
The compact jump model (1.7\,M parameters, $3\times$ smaller) recovers motion direction by re-anchoring on its own predictions, confirming that the architectural recipe is sound and the bottleneck is the training-inference alignment.
This finding provides a concrete path forward: per-step re-anchoring combined with stronger temporal supervision (e.g., scheduled sampling or perceptual losses) should enable recognizable multi-second predictions as scale increases.

\textbf{Compact scale as a feature.}
While large-scale systems like GAIA-1~\cite{hu2023gaia1} (5\,B parameters) and Cosmos~\cite{cosmos2024nvidia} demonstrate impressive generation quality, our compact-scale study ($\sim$5\,M parameters) offers complementary value: controlled ablations that isolate individual design factors are tractable at this scale but prohibitively expensive at 5\,B.
The four ingredients we identify (spatial tokens, $x_0$ objective, residual anchoring, sampling-matched-to-uncertainty) and the deployable calibration strategy are design principles that transfer to larger systems, and the positive scaling probe (Section~\ref{sec:pd}) provides initial evidence that they do.

\section{Conclusion}

We built a compact, action-conditioned DiT world model for AV scene prediction and, in the process, turned two obstacles into results.

First, distortion metrics that reward the blurry conditional mean masked the diffusion model's quality.
By measuring distribution realism via FID and KID, we revealed a clean perception-distortion frontier where diffusion is $4.8\times$ better on KID, made \emph{deployable} through a train-derived calibration.
This is a metric lesson for AV latent prediction: world models should be evaluated with distribution metrics, not distortion alone.

Second, single-pass rollouts produced sharp but temporally limited predictions.
We diagnosed the shared-anchor cause and showed that a compact jump model ($3\times$ smaller, re-anchored per step) recovers forward-motion \emph{direction} on held-out scenes, exceeding the larger baseline.
The model is also action-controllable ($\rho{=}0.81$), providing the planning-relevant property that a regression baseline lacks.

Together these establish a small but controllable AV world model with a characterized realism frontier and a validated route to coarse motion.

\textbf{Future work.}
The frontier and the jump result both point the same way: scale (capacity, data, frame rate) and stronger temporal supervision (e.g., perceptual losses, autoregressive scheduled sampling) to convert coarse motion \emph{direction} into recognizable high-fidelity appearance.
Closed-loop evaluation with predicted (rather than logged) actions and multi-camera setups are natural extensions.
Our controlled findings at compact scale provide a concrete recipe for the next iteration: the $x_0$ objective, residual anchoring, distribution metrics, deployable calibration, and the jump reparameterization.

\section*{Contributions and Acknowledgements}

\vspace{-1pt}
\textbf{Ruslan Sharifullin}: project direction and experiment planning, six-encoder benchmark design and probe training, DiT world model architecture, VAE encode-predict-decode pipeline, diffusion training and DDIM sampling, perception-distortion frontier analysis, FID/KID evaluation framework, action controllability experiments, motion fidelity diagnostics, chain-anchor jump model, cloud compute infrastructure, report writing and editing.

\textbf{Benjamin Jiang}: nuScenes data pipeline (frame/clip dataset, scene-level splits, CAN-bus extraction), RMSE evaluation harness with bootstrap CIs, perturbation analysis pipeline, environmental robustness analysis (night/rain/day bucketing), encoder attribution visualization (GradCAM), RMSE heatmaps and CosSim bar charts, report editing.

\textbf{Kai Xi Chew}: behavioral cloning baseline implementation and training, per-horizon CosSim and DeltaCosSim evaluation pipeline, embedding precomputation infrastructure, report editing.

\section*{Source Code}
\vspace{-6pt}
Source code, trained checkpoints and evaluation artifacts are available at \url{https://github.com/dlcv-team/latent-world-models-av}.

{\small
\bibliographystyle{ieee}
\bibliography{main}

\begin{thebibliography}{10}\itemsep=-1pt

\bibitem{diamond2024}
E.~Alonso, A.~Jelley, A.~Kanervisto, and T.~Pearce.
\newblock {DIAMOND}: Diffusion for world modeling.
\newblock {\em arXiv preprint arXiv:2405.12399}, 2024.

\bibitem{bardes2024vjepa}
A.~Bardes et~al.
\newblock {V-JEPA}: Latent video prediction for visual representation learning.
\newblock {\em arXiv preprint arXiv:2404.08471}, 2024.

\bibitem{bardes2025vjepa2}
A.~Bardes et~al.
\newblock {V-JEPA 2}: Self-supervised video models enable understanding,
  prediction and planning.
\newblock {\em arXiv preprint arXiv:2506.09985}, 2025.

\bibitem{binkowski2018kid}
M.~Bi{\'n}kowski, D.~J. Sutherland, M.~Arbel, and A.~Gretton.
\newblock Demystifying {MMD GANs}.
\newblock In {\em International Conference on Learning Representations (ICLR)},
  2018.

\bibitem{blau2018perception}
Y.~Blau and T.~Michaeli.
\newblock The perception-distortion tradeoff.
\newblock In {\em Proceedings of the IEEE Conference on Computer Vision and
  Pattern Recognition (CVPR)}, pages 6228--6237, 2018.

\bibitem{caesar2020nuscenes}
H.~Caesar et~al.
\newblock {nuScenes}: A multimodal dataset for autonomous driving.
\newblock In {\em IEEE/CVF Conference on Computer Vision and Pattern
  Recognition (CVPR)}, 2020.

\bibitem{dosovitskiy2021vit}
A.~Dosovitskiy et~al.
\newblock An image is worth 16x16 words: Transformers for image recognition at
  scale.
\newblock In {\em International Conference on Learning Representations (ICLR)},
  2021.

\bibitem{esser2021vqgan}
P.~Esser, R.~Rombach, and B.~Ommer.
\newblock Taming transformers for high-resolution image synthesis.
\newblock In {\em IEEE/CVF Conference on Computer Vision and Pattern
  Recognition (CVPR)}, 2021.

\bibitem{heusel2017fid}
M.~Heusel, H.~Ramsauer, T.~Unterthiner, B.~Nessler, and S.~Hochreiter.
\newblock {GANs} trained by a two time-scale update rule converge to a local
  {N}ash equilibrium.
\newblock In {\em Advances in Neural Information Processing Systems (NeurIPS)},
  2017.

\bibitem{ho2020ddpm}
J.~Ho, A.~Jain, and P.~Abbeel.
\newblock Denoising diffusion probabilistic models.
\newblock {\em Advances in Neural Information Processing Systems}, 2020.

\bibitem{ho2022classifier}
J.~Ho and T.~Salimans.
\newblock Classifier-free diffusion guidance.
\newblock {\em arXiv preprint arXiv:2207.12598}, 2022.

\bibitem{hu2023gaia1}
A.~Hu et~al.
\newblock {GAIA-1}: A generative world model with integrated action
  understanding.
\newblock {\em arXiv preprint arXiv:2309.17080}, 2023.

\bibitem{nichol2021improved}
A.~Q. Nichol and P.~Dhariwal.
\newblock Improved denoising diffusion probabilistic models.
\newblock {\em International Conference on Machine Learning (ICML)}, 2021.

\bibitem{cosmos2024nvidia}
{NVIDIA}.
\newblock Cosmos world foundation model platform for physical ai.
\newblock {\em arXiv preprint arXiv:2501.03575}, 2024.

\bibitem{oquab2024dinov2}
M.~Oquab et~al.
\newblock {DINOv2}: Learning robust visual features without supervision.
\newblock {\em Transactions on Machine Learning Research}, 2024.

\bibitem{peebles2023dit}
W.~Peebles and S.~Xie.
\newblock Scalable diffusion models with transformers.
\newblock In {\em International Conference on Computer Vision (ICCV)}, 2023.

\bibitem{polyak2024moviegen}
A.~Polyak et~al.
\newblock Movie gen: A cast of media foundation models.
\newblock {\em arXiv preprint arXiv:2410.13720}, 2024.

\bibitem{radford2021clip}
A.~Radford et~al.
\newblock Learning transferable visual models from natural language
  supervision.
\newblock In {\em International Conference on Machine Learning (ICML)}, 2021.

\bibitem{rombach2022ldm}
R.~Rombach, A.~Blattmann, D.~Lorenz, P.~Esser, and B.~Ommer.
\newblock High-resolution image synthesis with latent diffusion models.
\newblock In {\em IEEE/CVF Conference on Computer Vision and Pattern
  Recognition (CVPR)}, 2022.

\bibitem{shi2026drivewam}
C.~Shi, J.~Xu, S.~Shi, K.~Sheng, B.~Zhang, and L.~Jiang.
\newblock {DriveWAM}: Video generative priors enable scalable world-action
  modeling for autonomous driving.
\newblock {\em arXiv preprint arXiv:2605.28544}, 2026.

\bibitem{song2021ddim}
J.~Song, C.~Meng, and S.~Ermon.
\newblock Denoising diffusion implicit models.
\newblock In {\em International Conference on Learning Representations (ICLR)},
  2021.

\bibitem{tancik2020fourier}
M.~Tancik et~al.
\newblock Fourier features let networks learn high frequency functions in low
  dimensional domains.
\newblock In {\em Advances in Neural Information Processing Systems}, 2020.

\bibitem{yang2024driveworldvideo}
C.~Yang et~al.
\newblock Driveworld: 4d pre-trained scene understanding via world models for
  autonomous driving.
\newblock {\em arXiv preprint arXiv:2405.04390}, 2024.

\bibitem{yang2023unisim}
M.~Yang et~al.
\newblock {UniSim}: Learning interactive real-world simulators.
\newblock {\em arXiv preprint arXiv:2310.06114}, 2023.

\bibitem{zhao2024drivedreamer}
X.~Zhao et~al.
\newblock Drivedreamer: Towards real-world-driven world models for autonomous
  driving.
\newblock {\em arXiv preprint arXiv:2309.09777}, 2024.

\bibitem{zheng2025genad}
J.~Zheng et~al.
\newblock {GenAD}: Generalized predictive model for autonomous driving.
\newblock {\em arXiv preprint arXiv:2405.09349}, 2024.

\end{thebibliography}
}


\appendix
\clearpage
\onecolumn
\section{Additional Qualitative Results}

\begin{center}
\makeatletter
\includegraphics[height=0.78\textheight, angle=90]{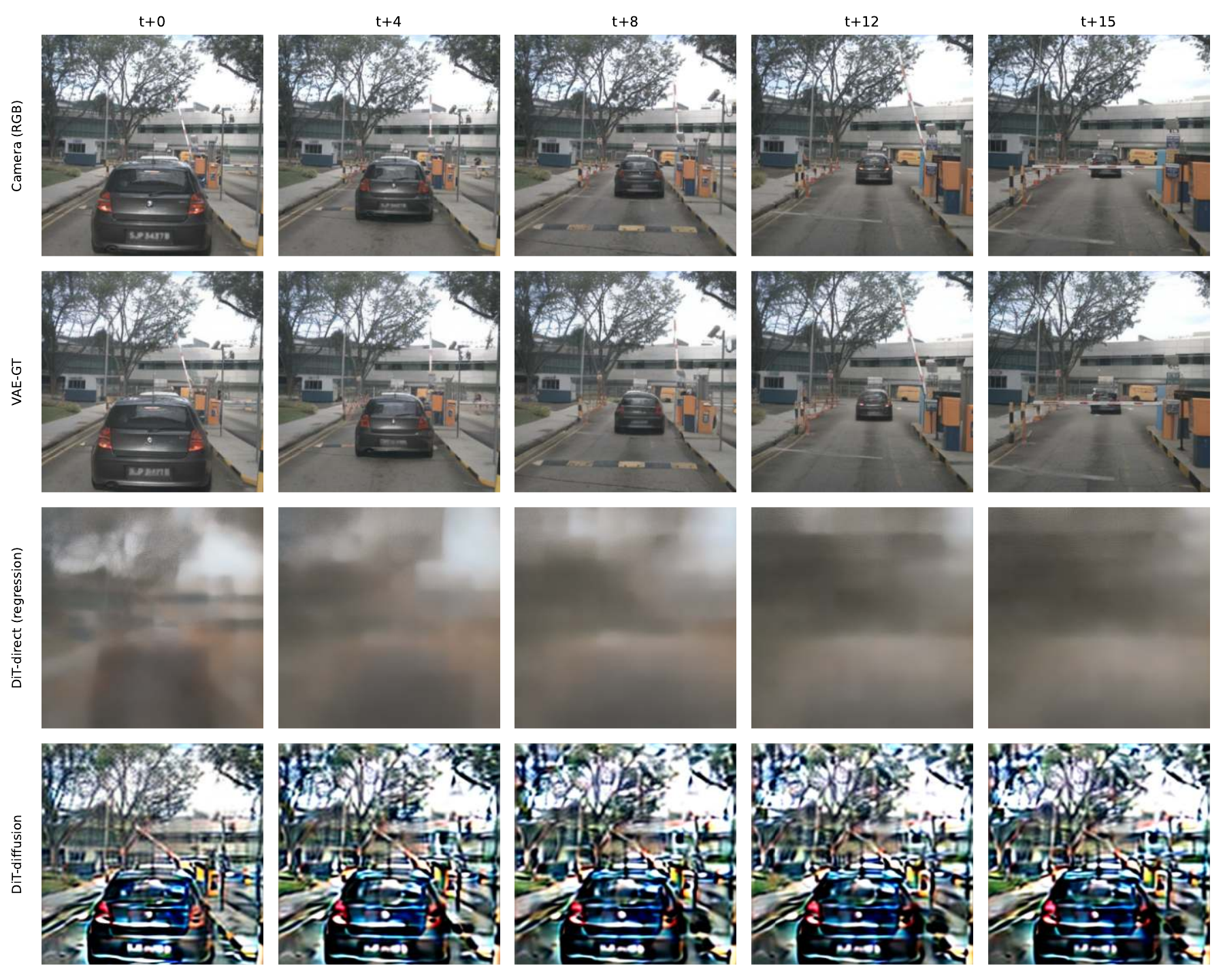}
\captionof{figure}{Qualitative VAE encode-predict-decode on a held-out nuScenes scene ($t{+}0$ through $t{+}15$).
Row~1: camera RGB ground truth.
Row~2: VAE-GT reconstruction (the appearance ceiling of this latent space).
Row~3: \textbf{DiT-direct prediction}, whose blur is regression \emph{mean-collapse} (not VAE tokenization loss).
Row~4: \textbf{DiT-diffusion prediction} (train-calibrated), which renders in-distribution frames with recognizable scene content.
This side-by-side contrast is the perception-distortion tradeoff made visual: the direct model minimizes distortion at the cost of sharpness, while the diffusion model produces realistic appearance at the cost of per-pixel fidelity.}
\label{fig:four_row}
\makeatother
\end{center}

\end{document}